\documentclass[runningheads]{llncs}
\usepackage[T1]{fontenc}
\usepackage{graphicx,wrapfig,lipsum,booktabs}
\usepackage{amsmath}
\usepackage{amssymb}
\usepackage{booktabs}
\usepackage{csquotes}
\usepackage{multirow}
\usepackage{mathtools}
\usepackage{makecell}
\usepackage[normalem]{ulem}
\usepackage[pagebackref,breaklinks,colorlinks]{hyperref}
\usepackage[usenames,dvipsnames,svgnames,table]{xcolor}
\usepackage{subcaption}
\usepackage{xspace}
\usepackage{amssymb, amsmath}
\usepackage{dsfont}
\usepackage{xcolor}

\makeatletter
\DeclareRobustCommand\onedot{\futurelet\@let@token\@onedot}
\def\@onedot{\ifx\@let@token.\else.\null\fi\xspace}

\makeatother

\renewcommand{\vec}[1]{{\mathbf #1}}
\usepackage{pifont}
\usepackage{hyperref}
\hypersetup{
    colorlinks=true,
    linkcolor=blue,
    filecolor=magenta,      
    urlcolor=cyan,
}
\usepackage[font=scriptsize]{caption}
\usepackage{mwe}
\usepackage[capitalize]{cleveref}
\crefname{section}{Sec.}{Secs.}
\Crefname{section}{Section}{Sections}
\Crefname{table}{Table}{Tables}

\begin{document}
\title{High-Resolution Swin Transformer for Automatic Medical Image Segmentation}
\titlerunning{HRSTNet}
\author{Chen Wei 
\inst{1}, Shenghan Ren 
\inst{2}, Kaitai Guo 
\inst{3}, Haihong Hu 
\inst{3}, Jimin Liang 
\inst{3}$^{\dagger}$}

\authorrunning{C. Wei et al.}
%
\institute{College of Economics and Management, Xi'an University of Posts\&Telecommunications, Xi'an, Shaanxi, China. \and School of Life Science and Technology, Xidian University, Xi’an, Shaanxi, China. \and School of Electronic Engineering, Xidian University, Xi’an, Shaanxi, China. \\
%
\email{weichen@xupt.edu.cn, \{shren, ktguo, hhhu, jimleung\}@mail.xidian.edu.cn} \\ $\dagger$ Corresponding author: Jimin Liang}

\maketitle              
\vspace{-8mm}
\begin{abstract}
The Resolution of feature maps are critical for medical image segmentation. Most of the existing Transformer-based networks for medical image segmentation are U-Net-like architecture that contains an encoder that utilizes a sequence of Transformer blocks to convert the input medical image from high-resolution representation into low-resolution feature maps and a decoder that gradually recovers the high-resolution representation from low-resolution feature maps. Unlike previous studies, in this paper, we utilize the network design style from the High-Resolution Network (HRNet), replace the convolutional layers with Transformer blocks, and continuously exchange information from the different resolution feature maps that are generated by Transformer blocks. The newly Transformer-based network presented in this paper is denoted as High-Resolution Swin Transformer Network (HRSTNet). Extensive experiments illustrate that HRSTNet can achieve comparable performance with the state-of-the-art Transformer-based U-Net-like architecture on Brain Tumor Segmentation(BraTS) 2021 and the liver dataset from Medical Segmentation Decathlon. The code of HRSTNet will be publicly available at \href{https://github.com/auroua/HRSTNet}{https://github.com/auroua/HRSTNet}.

\end{abstract}
\section{Introduction}
\label{sec:intro}
Although convolutional neural networks (CNNs) have been widely applied to different computer vision (CV) tasks, there still exist drawbacks of CNNs, such as the locality property of convolutional kernels preventing CNNs from providing a richer representation of contextual information for pixels. The performance of CV tasks may be compromised by the induction bias of CNNs, especially for dense prediction tasks. The Transformer, which has been successfully used in Natural Language Processing (NLP)~\cite{DBLP:conf/nips/BrownMRSKDNSSAA20,DBLP:conf/naacl/DevlinCLT19,DBLP:conf/acl/LewisLGGMLSZ20,Radford2018ImprovingLU,Radford2019LanguageMA,DBLP:journals/jmlr/RaffelSRLNMZLL20,DBLP:conf/nips/VaswaniSPUJGKP17}, excels at establishing long-range dependencies of different words in a sentence, which has attracted the attention of CV researchers. A collection of studies utilizing Transformer to improve the performance of image classification~\cite{DBLP:conf/iclr/DosovitskiyB0WZ21,DBLP:conf/icml/TouvronCDMSJ21,DBLP:conf/iccv/LiuL00W0LG21,DBLP:conf/iccv/WangX0FSLL0021}, object detection~\cite{Beal2020TowardTO,Mao2021VoxelTF,Misra2021AnET}, semantic segmentation~\cite{DBLP:conf/nips/XieWYAAL21,DBLP:conf/cvpr/ZhengLZZLWFFXT021}, representation learning~\cite{DBLP:conf/iccv/CaronTMJMBJ21,DBLP:conf/cvpr/Chen000DLMX0021,DBLP:conf/iccv/ChenXH21,He2021MaskedAA} and medical image segmentation~\cite{Hatamizadeh2022UNETRTF,Zhou2021nnFormerIT,10.1007/978-3-030-87199-4_16,10.1007/978-3-030-87193-2_11,DBLP:conf/miccai/PetitTRTCS21,DBLP:journals/corr/abs-2111-13300,DBLP:journals/corr/abs-2201-01266} have emerged in recent years.

Since a large number of medical images are generated each day, automatic medical image segmentation is urgently needed to aid doctors in the diagnosis of diseases. Existing medical image segmentation models~\cite{DBLP:conf/miccai/RonnebergerFB15,DBLP:conf/miccai/CicekALBR16,milletari2016v,Isensee2021} usually adopt U-Net-like architecture to generate semantically meaningful representations for segmentation by gathering local and global information for each pixel in the medical image. The U-Net-like architecture comprises an encoder and a decoder, the medical image passthrough the encoder and gradually generates spatially reduced and semantically richer low-resolution representations which contain global contextual information, and the decoder takes the low-resolution representations as input and passes them through several network layers to recover the high-resolution representations for segmentation. To further improve the global contextual information of representations, researchers replace CNN layers in the U-Net-like architecture with Transformer, and the newly generated Transformer-based architectures~\cite{Zhou2021nnFormerIT,DBLP:journals/corr/abs-2111-13300,DBLP:journals/corr/abs-2201-01266} achieve state-of-the-art (SOTA) performance on several medical image benchmarks.

Although U-Net-like architecture can provide affluent contextual information, the procedure of recovering high-resolution representations from low-resolu-tion representations may harm the spatial precision of the generated segmentation masks. The HRNet~\cite{9052469} maintains the high-resolution representations and connects the high-to-low resolution representations in parallel, which can remove the necessity to recover high-resolution representations from low-resolution representations and hence may generate more spatially precision representations, Unlike existing Transformer-based methods for medical image segmentation that all adopt the U-Net-like architecture, in this paper, we propose the High-Resolution Swin Transformer Network (HRSTNet) that combines Transformer with the HRNet architecture for volumetric medical image segmentation. Concretely, we employ the Swin Transfomer block~\cite{DBLP:conf/iccv/LiuL00W0LG21} to generate parallel feature representations, utilize patch merging and expanding blocks to downsample and upsample feature representations and design a multi-resolution feature fusion block to fusion features of different resolutions. 

In summary, our main contributions are, \textbf{(1)} We propose the HRSTNet that combines HRNet with the Swin Transformer block, the Swin Transformer blocks generate parallel features of different resolutions, and the high-resolution features are maintained throughout the network to provide spatially more precise information, while the multi-resolution feature fusion block is utilized to generate contextual information augmented representations. \textbf{(2)} The multi-resolution feature fusion block is designed to fusion features of different resolutions, and it utilizes the patch merging block to downsample feature maps' resolution and the patch expanding block to upsample feature maps' resolution. \textbf{(3)} By recovering the segmentation mask from different stages of HRSTNet, a series of networks with different FLOPs are designed for different scenarios. \textbf{(4)} Extensive experiments on BraTS 2021~\cite{Baid2021TheRB,menze2014multimodal,bakas2017segmentation,bakas2017segmentation1,bakas2017advancing} and the liver dataset from Medical Segmentation Decathlon~\cite{antonelli2021medical,simpson2019large} show that the HRSTNet can achieve comparable or even better performance with recently proposed medical image segmentation methods.

\section{Related Works}
\label{sec:literature}

\subsection{Vision Transformers}
Inspired by the successful use of Transformer in natural language processing tasks, ViT~\cite{DBLP:conf/iclr/DosovitskiyB0WZ21} splits the input image into $16 \times 16$ patches and converts the patches into tokens. After converting, an additional learnable classification token together with the image tokens passthrough the Transfomer encoder, and the generated features corresponding to the classification token are used to perform the classification task. By pre-training ViT on a large dataset that contains 303M high-resolution images and fine-tuning to a smaller dataset, ViT can achieve comparable performance with CNNs. DeiT~\cite{DBLP:conf/icml/TouvronCDMSJ21} designs a distillation method and a collection of training strategies that enables ViT to achieve comparable performance with CNNs without the need for pre-training on a large dataset. Swin Transformer~\cite{DBLP:conf/iccv/LiuL00W0LG21} divides the patches into non-overlapping windows and restricts the self-attention calculation in the small window or shift window, which introduces locality inductive bias. It also utilizes the patch merging layers to generate a hierarchical representation which can benefit the downstream tasks such as object detection or semantic segmentation. HRFormer~\cite{DBLP:conf/nips/YuanFHLZCW21} and HRViT~\cite{DBLP:journals/corr/abs-2111-01236} borrow the network architecture from HRNet~\cite{9052469} and design a Transformer-based architecture that maintains parallel resolution features and exchanges information from different resolutions. HRFormer utilizes local-window self-attention to generate features of different resolutions, and the features may lack contextual information from other windows. HRViT proposes the HRViTAttn block to generate different resolution features and the MixCFN block to exchange information from features of different resolution. Unlike HRFormer and HRViT, which are designed for 2D image segmentation tasks, we utilize the Swin Transformer block to create an HRNet-like architecture and apply the architecture to volumetric medical image segmentation tasks.

\subsection{Medical Image Segmentation}
Since the excellent performance of U-Net architecture~\cite{DBLP:conf/miccai/RonnebergerFB15} for medical image segmentation, recently proposed Transformer-based medical image segmentation methods~\cite{Hatamizadeh2022UNETRTF,Zhou2021nnFormerIT,10.1007/978-3-030-87199-4_16,10.1007/978-3-030-87193-2_11,DBLP:conf/miccai/PetitTRTCS21,DBLP:journals/corr/abs-2111-13300,DBLP:journals/corr/abs-2201-01266} utilize the U-Net-like architecture to gather the global and local contextual information for each pixel. UNETR~\cite{Hatamizadeh2022UNETRTF} adopts ViT as an encoder and converts the features from four different stage features of ViT to hierarchical feature maps by using a collection of deconvolutional layers, and the hierarchical feature maps are fed into the decoder to generate the segmentation mask. The nnFormer~\cite{Zhou2021nnFormerIT} contains a Transformer-based encoder, decoder, and bottleneck blocks, the encoder and decoder focus on extracting local information by using the local self-attention layer, while the bottleneck block is in charge of gathering global contextual information. TransBTS~\cite{10.1007/978-3-030-87193-2_11}, CoTr~\cite{10.1007/978-3-030-87199-4_16} and U-Transformer~\cite{DBLP:conf/miccai/PetitTRTCS21} adopt a CNN-encoder to extract features and utilize the Transformer layer to operate on the CNN features. TransBTS utilizes a standard Transformer layer, while CoTr designs a DeTrans layer that pays attention to a small set of key positions to reduce its computational cost. VT-UNet~\cite{DBLP:journals/corr/abs-2111-13300} adopts the Swin Transformer block to generate the hierarchical representation features and proposes the VT decoder block to merge the features from the encoder and decoder. Swin UNETR~\cite{DBLP:journals/corr/abs-2201-01266} designs the architecture like UNETR but utilizes the Swin Transformer as encoder. Like VT-UNet and Swin UNETR, we adopt the Swin Transformer block to extract features, but in contrast to the recently proposed Transformer-based medical image segmentation methods, we follow the HRNet architecture and maintain the high-resolution features to provide spatially precise information.
\section{Methodology}
\label{sec:method}

Inspired by HRNet, HRSTNet maintains a high-resolution representation and continually exchanges information from different resolution features. We delve into the details of the spatially more precise and contextually informative HRSTNet in this section.

\begin{figure*}[h]
    \scriptsize
    \centering
    \includegraphics[width=12cm, height=5.12cm]{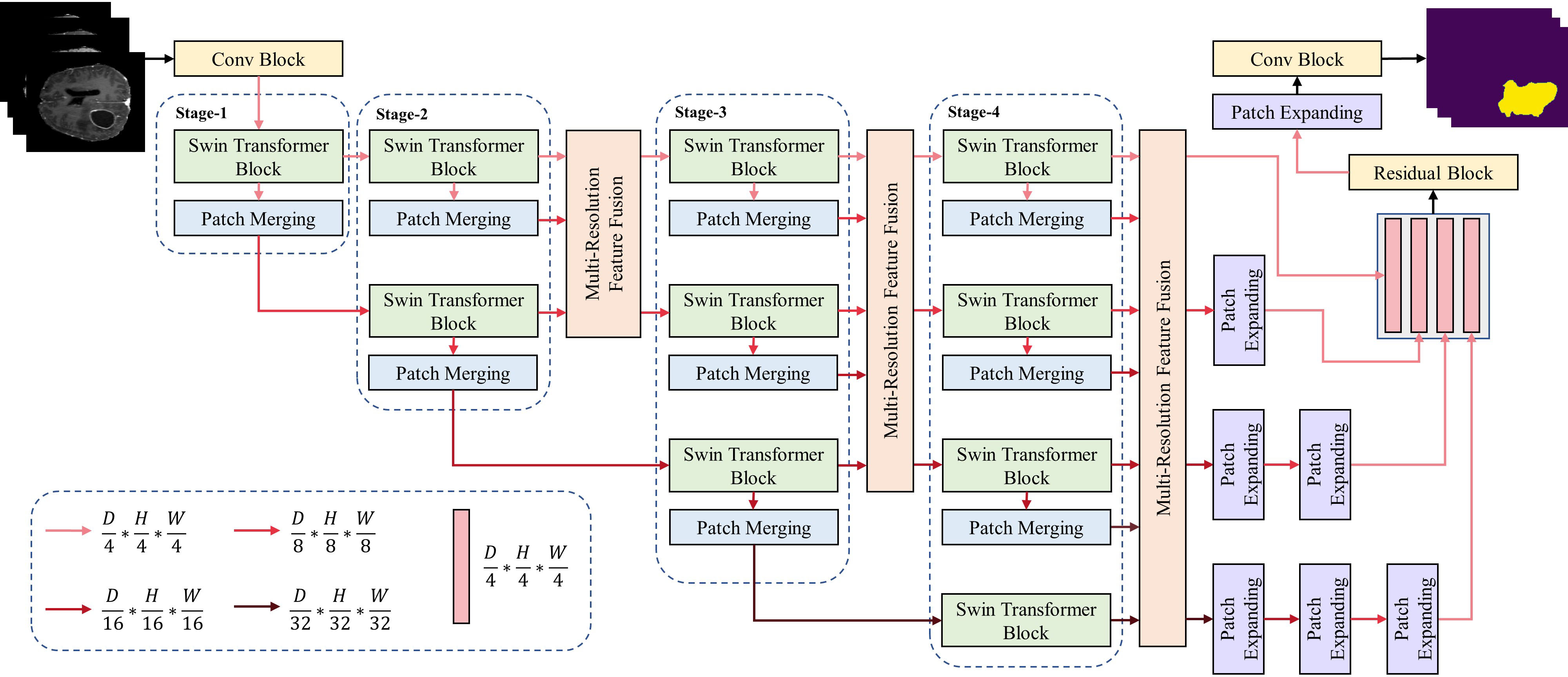}
    \caption{Overview of HRSTNet.}
    \label{fig:architecture}
\end{figure*}

\subsection{Architecture Overview}
As illustrated in Fig. \ref{fig:architecture}, HRSTNet contains several sequentially connected stages with multi-resolution feature fusion module insert into the stages (except for the first stage). The high-resolution features hold constant in all stages, and the \textit{n}-th stage contains \textit{n parallel} Swin Transformer blocks that will produce \textit{n parallel} semantically richer low-resolution features. The input 3D medical image feed into a 3D convolutional block (Conv block) that splits the 3D medical image into small 3D patches and represents each 3D patch as a vector in the feature space. The 3D patches' vectors pass through several stages, the outputs of the last stage are fed into the last multi-resolution feature fusion block, and the outputs of the above feature fusion block are concatenated after upsampling low-resolution features. The final segmentation mask is generated by passing the concatenated feature through a residual block, a patch expanding block, and a Conv block. The patch merging block and the patch expanding block are used to downsampling and upsampling the feature maps, and the definition of the patch merging block is the same as the Swin Transformer~\cite{DBLP:conf/iccv/LiuL00W0LG21}, while the patch expanding block is the same as VT-Unet~\cite{DBLP:journals/corr/abs-2111-13300}. The detailed information of the Swin Transformer block, HRSTNet stages and the multi-resolution feature fusion (MRFF) block are introduceed in the following subsections.

\begin{figure*}[h]
    \scriptsize
    \centering
    \includegraphics[width=12cm, height=6.857cm]{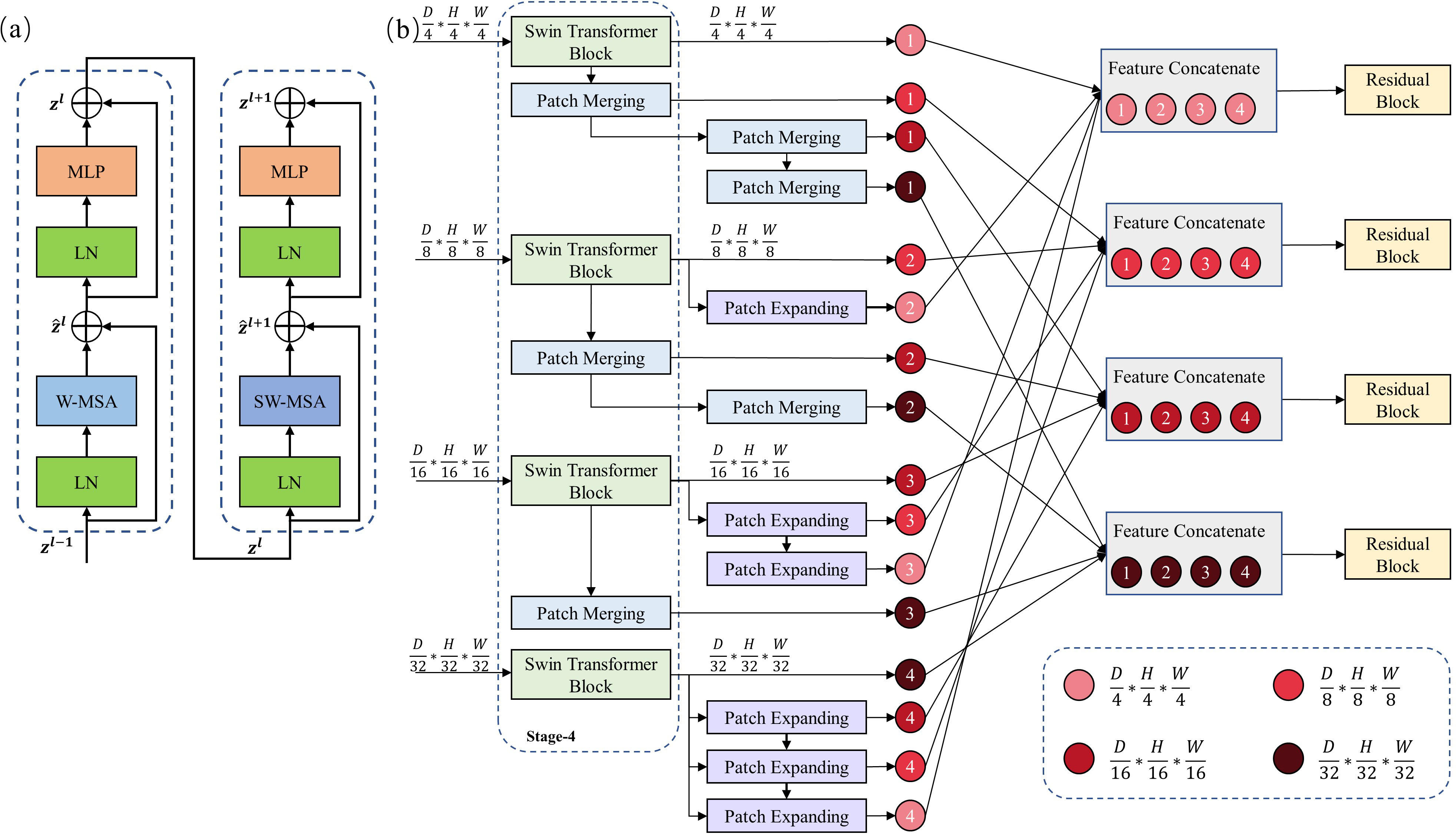}
    \caption{(a) shows the Swin Transformer block. (b) illustrates the multi-resolution feature fusion block.}
    \label{fig:swin_ff}
\end{figure*}

\subsection{Swin Transformer Block}
The Swin Transformer block is introduced in the previous studies~\cite{DBLP:conf/iccv/LiuL00W0LG21,DBLP:journals/corr/abs-2106-13230}, and it is illustrated in Fig. \ref{fig:swin_ff}(a). The Swin Transformer block contains two cascading layers, the first layer calculates the self-attention in a small window (W-MSA) of the input 3D patches and the second layer calculates the self-attention in the shifted window (SW-MSA) of the outputs of the previous layer.

Assuming the input 3D medical image is $X \in \mathbb{R}^{D \times H \times W \times K}$, where $K$ is the number of channels of the input 3D data. The first step of the Swin Transformer is to split $X$ into small patches, the patch size is set to $P \times P \times P$, so there are $S = \lfloor \frac{D}{P} \rfloor \times \lfloor \frac{H}{P} \rfloor \times \lfloor \frac{W}{P} \rfloor$ patches. Let $\mathds{X} = \{\vec{x}_1,\vec{x}_2,\cdots,\vec{x}_i,\cdots,\vec{x}_S\}, \vec{x_i} \in \mathbb{R}^{P \times P \times P \times K}$ denotes the collection of 3D patches. As illustrated in Fig. \ref{fig:architecture}, the collection of patches $\mathds{X}$ are fed into the Conv block, the kernel size and stride size of the Conv block are set equal to the patch size, and the output of the Conv block is denoted as $Z \in \mathbb{R}^{\lfloor \frac{D}{P} \rfloor \times \lfloor \frac{H}{P} \rfloor \times \lfloor \frac{W}{P} \rfloor \times C}$. Since each pixel in $Z$ represents the patch embedding vector, we resize the feature map $Z$ into a 2D matrice $\vec{Z} \in \mathbb{R} ^ {S \times C}$, where each row of the matrice $\vec{Z}$ represents a patch embedding vector, and there are $s$ rows in total. After patch embedding, the set of patch embedding vectors is fed into the Swin Transformer block, and the output of the Swin Transformer block is calculated as follows:

\begin{align}
    &{{\hat{\bf{Z}}}^{l}} = \text{W-MSA}\left( {\text{LN}\left( {{{\bf{Z}}^{l - 1}}} \right)} \right) + {\bf{Z}}^{l - 1}  \\
    &{{\bf{Z}}^l} = \text{MLP}\left( {\text{LN}\left( {{{\hat{\bf{Z}}}^{l}}} \right)} \right) + {{\hat{\bf{Z}}}^{l}} \\
    &{{\hat{\bf{Z}}}^{l+1}} = \text{SW-MSA}\left( {\text{LN}\left( {{{\bf{Z}}^{l}}} \right)} \right) + {\bf{Z}}^{l} \\
    &{{\bf{Z}}^{l+1}} = \text{MLP}\left( {\text{LN}\left( {{{\hat{\bf{Z}}}^{l+1}}} \right)} \right) + {{\hat{\bf{Z}}}^{l+1}},
    \label{eqn:swin_block}
\end{align}
where the definition of W-MSA, SW-MSA, LN, and MLP are the same as in Swin Transformer~\cite{DBLP:conf/iccv/LiuL00W0LG21}.

\subsection{HRSTNet Stage}
The HRSTNet contains several stages, and each stage will generate multi-resolu-tion feature maps, except for the last stage, the $n$-th stage contains $n$ Swin Transformer blocks and $n$ patch merging blocks, while the last stage contains $n-1$ patch merging blocks. In the $n$-th stage, the $n$ parallel Swin Transformer blocks process $n$ different resolution feature maps, respectively. Each Swin Transformer block is followed by a patch merge, which is used to downsample the output feature map of the Swin Transformer block.

Take the third stage (Stage-3) as an example, the input to this stage is three feature maps with resolutions of $\frac{D}{4} \times \frac{H}{4} \times \frac{W}{4}$, $\frac{D}{8} \times \frac{H}{8} \times \frac{W}{8}$, and $\frac{D}{16} \times \frac{H}{16} \times \frac{W}{16}$, respectively. The three Swin Transformer blocks process the three feature maps, and the output of the Swin Transformer block has the same resolution as its input feature map. The output of the Swin Transformer is fed into a patch merging layer, thus yielding three feature maps with resolutions of $\frac{D}{8} \times \frac{H}{8} \times \frac{W}{8}$, $\frac{D}{16} \times \frac{H}{16} \times \frac{W}{16}$, and $\frac{D}{32} \times \frac{H}{32} \times \frac{W}{32}$. The feature map with resolution $\frac{D}{32} \times \frac{H}{32} \times \frac{W}{32}$ is directly fed into the fourth stage (Stage-4), while the outputs of other patch merging layers and the outputs of the Swin Transformer blocks are fed into the multi-resolution feature fusion block. 

\subsection{Multi-Resolution Feature Fusion}
As shown in Fig. \ref{fig:swin_ff}(b), the multi-resolution feature fusion block is used to exchange information from feature maps with different resolutions and produce new feature maps that are spatially more precise and semantically richer. Fig. \ref{fig:swin_ff}(b) shows the MRFF block that follows the fourth stage, and this feature fusion block fuse features of four different resolutions. 

The fourth stage of HRSTNet contains four Swin Transformer blocks, which process four different resolutions of feature maps, and there are four feature maps with resolutions $\frac{D}{4} \times \frac{H}{4} \times \frac{W}{4}$, $\frac{D}{8} \times \frac{H}{8} \times \frac{W}{8}$, $\frac{D}{16} \times \frac{H}{16} \times \frac{W}{16}$, and $\frac{D}{32} \times \frac{H}{32} \times \frac{W}{32}$ generated by the four Swin Transformer blocks. The feature maps with resolutions $\frac{D}{4} \times \frac{H}{4} \times \frac{W}{4}$, $\frac{D}{8} \times \frac{H}{8} \times \frac{W}{8}$, $\frac{D}{16} \times \frac{H}{16} \times \frac{W}{16}$, and $\frac{D}{32} \times \frac{H}{32} \times \frac{W}{32}$ are referred to as $f_4, f_8, f_{16}$, and $f_{32}$, respectively. Each of the four feature maps will be downsampled or upsampled to the other three resolutions.
Take feature map $f_4$ as an example, $f_4$ sequentially passes through three patch merging layers to generate three new feature maps with resolutions of $\frac{D}{8} \times \frac{H}{8} \times \frac{W}{8}$, $\frac{D}{16} \times \frac{H}{16} \times \frac{W}{16}$, and $\frac{D}{32} \times \frac{H}{32} \times \frac{W}{32}$, and the three newly generated feature maps are denoted as $f_{4\_8}$, $f_{4\_16}$, and $f_{4\_32}$. It can be easily infered the feature maps generated by $f_8$, $f_{16}$, and $f_{32}$, and the generated feature maps are denoted as $\{f_{8\_4}, f_{8\_16}, f_{8\_32}\}$, $\{f_{16\_4}, f_{16\_8}, f_{16\_32}\}$, and $\{f_{32\_4}, f_{32\_8}, f_{32\_16}\}$, respectively. We then concatenate feature maps with the same resolution, e.g., $ [f_{4}, f_{8\_4}, f_{16\_4}, f_{32\_4}]$, where the symbol $[ \quad ]$ indicates feature concatenation. After concatenating the feature maps, the concatenated feature maps are fed to the residual block to produce the output of the MRFF block, thus producing four feature maps with resolutions of $\frac{D}{4} \times \frac{H}{4} \times \frac{W}{4}$, $\frac{D}{8} \times \frac{H}{8} \times \frac{W}{8}$, $\frac{D}{16} \times \frac{H}{16} \times \frac{W}{16}$, and $\frac{D}{32} \times \frac{H}{32} \times \frac{W}{32}$. The output of the MRFF blocks after the second and third stages can be easily derived. As shown in Fig. \ref{fig:architecture}, the outputs of the last MRFF block with a resolution lower than $\frac{D}{4} \times \frac{H}{4} \times \frac{W}{4}$ are upsampled. After upsampling, all the feature maps are concatenated and used to generate the output segmentation mask.

Except for using the last MRFF block shown in Fig. \ref{fig:architecture} to generate the output segmentation mask, we can also use the outputs of the MRFF block that follows the second stage or follows the third stage to generate the segmentation mask. The architectures that only use the outputs of the MRFF block that follows the second and third stages to generate the segmentation mask are denoted as HRSTNet-2 and HRSTNet-3, respectively. The architecture shown in Fig. \ref{fig:architecture} is denoted as HRSTNet-4.
\section{Experiments}
\label{sec:exp}
In this section, we illustrate the performance of HRSTNet on the BraTS 2021 dataset~\cite{Baid2021TheRB,menze2014multimodal,bakas2017segmentation,bakas2017segmentation1,bakas2017advancing} and the liver dataset from MSD~\cite{antonelli2021medical,simpson2019large}. All the experiments are implemented in Pytorch~\cite{Paszke2019PyTorchAI}. We utilize MONAI\footnote[1]{https://monai.io/}, a healthcare imaging processing framework, for network training and data pre-processing. 

\subsection{Implementation Details}
The BraTS 2021 contains 1251 MRI scans with shape $240\times 240 \times 155$, and following VT-UNet~\cite{DBLP:journals/corr/abs-2111-13300}, the 1251 MRI scans are split into 834, 208, and 209 for training, validation, and testing, respectively. The task is to segment the three semantically meaningful tumor classes, namely enhanced tumor (ET), tumor core (TC) region, and whole tumor (WT) region. The liver dataset from MSD contains 131 CT volumes. The 131 liver CT volumes are split into 87, 22, and 22 for training, validation, and testing, respectively. The tasks for the liver datasets are tumor segmentation.

The depth of Swin Transformer block in each stage is two, and the heads of Swin Transformer blocks for resolutions $\frac{D}{4} \times \frac{H}{4} \times \frac{W}{4}$, $\frac{D}{8} \times \frac{H}{8} \times \frac{W}{8}$, $\frac{D}{16} \times \frac{H}{16} \times \frac{W}{16}$, and $\frac{D}{32} \times \frac{H}{32} \times \frac{W}{32}$ are 3, 6, 12, and 24, respectively. The AdamW optimizer is used to optimize the parameters of HRSTNet, and the learning rate is $1e^{-4}$. The cosine decay learning rate scheduler~\cite{Loshchilov2017SGDRSG} with linear warmup is used to adjust the value of the learning rate, and the warmup epoch is 50. The sum of the dice loss with the cross-entropy loss is adopted as the loss function. For a fair comparison, on the BraTS 2021 dataset, the dataset preprocessing and the evaluation method of models are the same as VT-UNet~\cite{DBLP:journals/corr/abs-2111-13300}. The batch size is 1, the BraTS MRI scans are cropped to a fixed size of $128 \times 128 \times 128$, and the cropped MRI scans are fed to the HRSTNet. To illustrate its superior performance, on the BraTS 2021 dataset, the HRSTNet is compared with three Transformer-based methods, UNETR~\cite{Hatamizadeh2022UNETRTF}, VT-UNet~\cite{DBLP:journals/corr/abs-2111-13300}, and Swin UNETR~\cite{DBLP:journals/corr/abs-2201-01266}. We train the above methods for 300 epochs on the training dataset of BraTS 2021, and compare the performance of these methods on the testing dataset of BraTS 2021. 

The CT volumes from the liver dataset are cropped to $96\times96\times96$. Other training details are the same as training on the BraTS 2021 dataset.Since the liver dataset splitting method in this paper is the same as VT-UNet, we compare the HRSTNet with the methods listed in VT-UNet. The compared methods are 3D UNet~\cite{DBLP:conf/miccai/CicekALBR16}, nnFormer~\cite{Zhou2021nnFormerIT}, and VT-UNet~\cite{DBLP:journals/corr/abs-2111-13300}, and the experiment results are employed from the VT-UNet directly.

The model with the best Dice-Similarity coefficient (DSC) on the validation dataset is saved for final evaluation, and model evaluation tricks such as model ensembles are not used in this paper. All the experiments in this section are performed by using two RTX 3090 GPUs. For a fair comparison, the three compared methods are retrained by using the same dataset preprocessing methods, training devices, and evaluation method, and the results of the retraining are reported.

\subsection{Experimental Results}
The Dice score and Hausdorff Distance (HD) are used to quantitatively evaluate the segmentation result on the BraTS 2021 dataset, while the Dice score is utilized on the liver dataset from MSD. For a fair comparison, we utilize the method provided by nnUNet\footnote[2]{https://github.com/MIC\-DKFZ/nnUNet/blob/5c18fa32f2b31575aae59d889d196e4c\\4ba8b844/nnunet/dataset\_conversion/Task082\_BraTS\_2020.py\#L330} to calculate the Hausdorff distance. 

As illustrated in Table \ref{tab:quantitative_results_BraTS} for the experimental results of the BraTS 2021 dataset, our proposed HRSTNet-4 achieves the best average Dice score, while HRSTNet-3 obtains the smallest average Hausdorff distance. Even the method HRSTNet-2, which has the lowest FLOPs among the series of HRSTNet, has a better average Hausdorff distance than the other three Transformer-based methods, and its average Dice score is comparable to that of VT-UNet. For enhancing tumor segmentation, the Dice score of VT-UNet-B performs the best, but its Hausdorff distance is the largest of all compared methods. Swin UNETR has a comparable Dice score with UNETR, but its Hausdorff score is better than UNETR. Considering that the difference between these two networks is the encoder, we conjecture that the ability of the Transformer to construct the long-term dependency leads to the improvement. Overall, the experimental results illustrated in Table \ref{tab:quantitative_results_BraTS} indicate that the HRNet-like network can achieve better performance than the UNet-like architectures, which implies that HRNet-like network design is a valuable direction to dive into.

\begin{minipage}[c]{0.85\textwidth}
\centering
\resizebox{1\textwidth}{!}{
\begin{tabular}{l|cc|cc|cc|cc}
    \toprule
       \multirow{2}{*}{Methods}  & \multicolumn{2}{c|}{Average} & \multicolumn{2}{c|}{WT} & \multicolumn{2}{c|}{ET} & \multicolumn{2}{c}{TC}\\  \cline{2-9}
       &  HD95 $\downarrow$ & DSC $\uparrow$ & HD95 $\downarrow$ & DSC $\uparrow$ & HD95 $\downarrow$ & DSC $\uparrow$ & HD95 $\downarrow$ & DSC $\uparrow$ \\
       \hline
       \hline
       UNETR~\cite{Hatamizadeh2022UNETRTF} & 11.04 & 86.18 & 6.18 & 91.40 & 18.31 & 81.80 & 8.63 & 85.33 \\
       Swin UNETR~\cite{DBLP:journals/corr/abs-2201-01266} & 9.84 & 86.13 & 6.06 & 90.42 & 15.27 & 83.36 & 8.19 & 84.62 \\
       VT-UNet-B~\cite{DBLP:journals/corr/abs-2111-13300} & 13.36 & 87.31 & 8.96 & 91.20 & 18.69 & \textbf{83.85} & 12.44 & 86.87 \\
       \hline
       HRSTNet-2 & 8.99 & 87.22 & 4.15 & \textbf{92.06} & 15.63 & 82.64 & 7.20 & 86.95 \\
       HRSTNet-3 & \textbf{8.06} & 86.94 & 4.79 & 91.81 & \textbf{13.91} & 82.20 & \textbf{5.50} & 86.80 \\
       HRSTNet-4 & 8.94 & \textbf{87.48} & \textbf{4.09} & 91.90 & 15.62 & 82.92 & 7.11 & \textbf{87.62} \\
    \bottomrule
\end{tabular}
}
\captionof{table}{Segmentation Results on BraTS 2021.}
\label{tab:quantitative_results_BraTS}
\end{minipage}

\begin{figure*}[hb!]
    \scriptsize
    \centering
    \includegraphics[width=12.5cm, height=6.56cm]{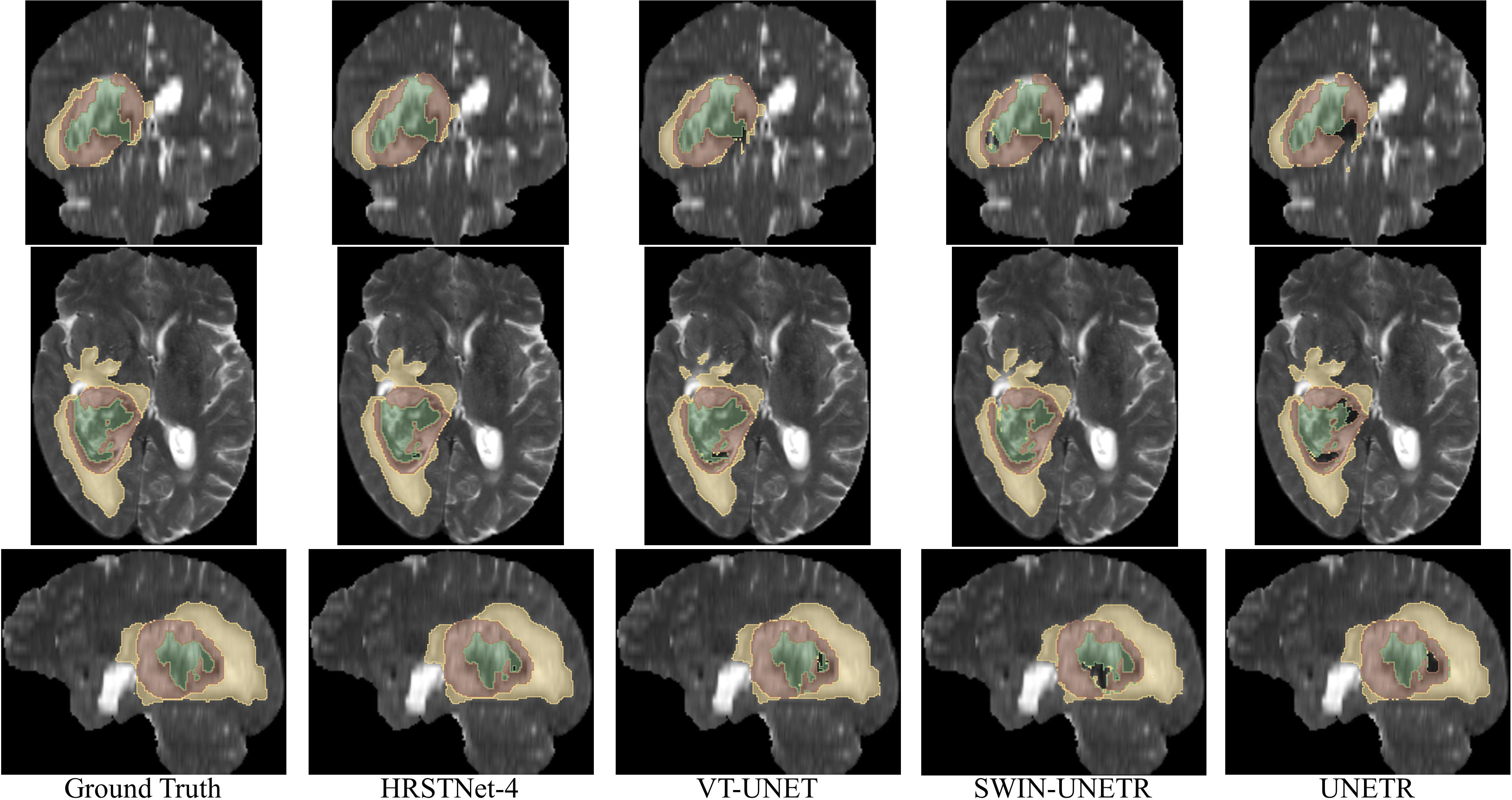}
    \caption{Visualizing of the segmentation results on BraTS 2021. Row 1 shows the coronal view of the segmentation results, row 2 illustrates the axial view of the segmentation results, and row 3 demonstrates the segmentation results of the sagittal view. The colors of the green, brown, and yellow represents the necrotic tumor core (NCR), enhancing tumor (ET), and the peritumoral edematous (ED), respectively.}
    \label{fig:visualization}
\end{figure*}

Fig. \ref{fig:visualization} shows the visualization of the segmentation results on BraTS 2021, from which we visually find that the segmentation results of HRSTNet-4 are better than those of the Transformer-based UNet-like architecture.

As demonstrated in Table \ref{tab:quantitative_results_MSD}, the performance of HRSTNet-4 surpasses the previous methods by a large margin of $+9.63$ tumor Dice score and $+3.76$ average Dice score for the liver tumor segmentation. The experimental results demonstrate that besides the segmentation of MRI scans, HRSTNet is also excellent in CT volume segmentation. Again, we clarify that no model ensemble is used in the experiments conducted in this section.
\begin{center}
    \resizebox{0.55\textwidth}{!}{
        \begin{tabular}{l|ccc}
        \hline
        \multirow{2}{*}{Methods}  & \multicolumn{3}{c}{Liver}  \\  \cline{2-4}
        &  Organ & Tumor & AVG.  \\
        \hline
        \hline
        3D UNet~\cite{DBLP:conf/miccai/CicekALBR16} & 92.67 & 34.92 & 63.80  \\
        nnFormer~\cite{Zhou2021nnFormerIT} & 89.43 & 31.84 & 60.63 \\
        VT-UNet-B~\cite{DBLP:journals/corr/abs-2111-13300} & \textbf{92.84} & 35.69 & 64.26 \\
        \hline
        HRSTNet-4 & 90.72 & \textbf{45.32} & \textbf{68.02} \\
        \hline
        \end{tabular}
    }
        \captionof{table}{Segmentation Results on the liver dataset from MSD.}
        \label{tab:quantitative_results_MSD}
\end{center}

\section{Conclusion}
In this paper, we present an HRNet-like architecture HRSTNet for medical image segmentation, and experimental results illustrate the superior performance of HRSTNet. For future studies, we will conduct experiments on more datasets to verify the performance of HRSTNet, and we will also design new HRNet-like that has better performance for medical image segmentation.
%
%
\bibliographystyle{ieeetr}
\bibliography{references}

\clearpage

\end{document}